\documentclass[letterpaper, 10 pt, conference]{ieeeconf}

\IEEEoverridecommandlockouts
\usepackage[table]{xcolor}
\usepackage{amsmath}
\usepackage{amssymb}
\usepackage{algorithm}
\usepackage{algpseudocode}
\usepackage{booktabs}
\usepackage{multirow}
\usepackage{tabularx}
\usepackage{graphicx}
\usepackage{url}
\usepackage{etoolbox}
\definecolor{deepblue}{RGB}{33,95,154}
\usepackage[colorlinks=true,
            citecolor=deepblue,
            linkcolor=deepblue,
            urlcolor=black]{hyperref}
\usepackage{cuted}
\usepackage{capt-of}
\usepackage{wrapfig}

\algrenewcommand\algorithmicrequire{\textbf{Input:}}
\algrenewcommand\algorithmicensure{\textbf{Output:}}

\algrenewcommand\algorithmiccomment[1]{%
    \hfill{\scriptsize\itshape #1}}

\newcommand{\AlgStage}[1]{%
    \Statex \vspace{-0.4em}
    \Statex {\color{deepblue}\textit{#1}}
}

\makeatletter
\patchcmd{\@makecaption}
  {\begin{center}{\footnotesize #1}\\{\footnotesize\scshape #2}\end{center}}
  {\begin{center}{\footnotesize\upshape #1: #2}\end{center}}{}
  {\PackageError{caption-migration}{Table caption patch failed}{Check the supplied ieeeconf class.}}
\makeatother

\title{\LARGE \bf
PaCo-VLA: \underline{Pa}ssivity-Shielded \underline{Co}mpliance Prior for \\ Contact-Rich Vision-Language-Action Manipulation
}

\author{
Haofan Cao$^{1,2}$\quad
Zhaoyang Li$^{1}$\quad
Zhichao You$^{1,*}$ \\
$^{1}$Southwest Jiaotong University, $^{2}$University of Leeds
\thanks{*Corresponding author.}%
}

\newcommand{\wrenchk}{\mathcal{F}_k}
\newcommand{\xk}{\mathbf{x}_k}
\newcommand{\vk}{\mathbf{v}_k}
\newcommand{\zk}{\mathbf{z}_k}

\newcommand{\methodname}{\textbf{PaCo-VLA}}
\newcommand{\best}[1]{\cellcolor{keycell}\textbf{#1}}
\newcommand{\bad}[1]{\cellcolor{badcell}#1}
\newcommand{\tracepaco}[1]{\textcolor{tracepaco}{\textbf{#1}}}
\newcommand{\tracevanilla}[1]{\textcolor{tracevanilla}{\textbf{#1}}}
\definecolor{oursrow}{RGB}{232,242,255}
\definecolor{keycell}{RGB}{181,212,249}
\definecolor{tracepaco}{RGB}{47,158,101}
\definecolor{tracevanilla}{RGB}{190,55,50}
\definecolor{badcell}{RGB}{255,198,198}

\begin{document}
\maketitle
\thispagestyle{empty}
\pagestyle{empty}

\begin{strip}
\centering
\vspace{-3.5em}
\includegraphics[width=\linewidth]{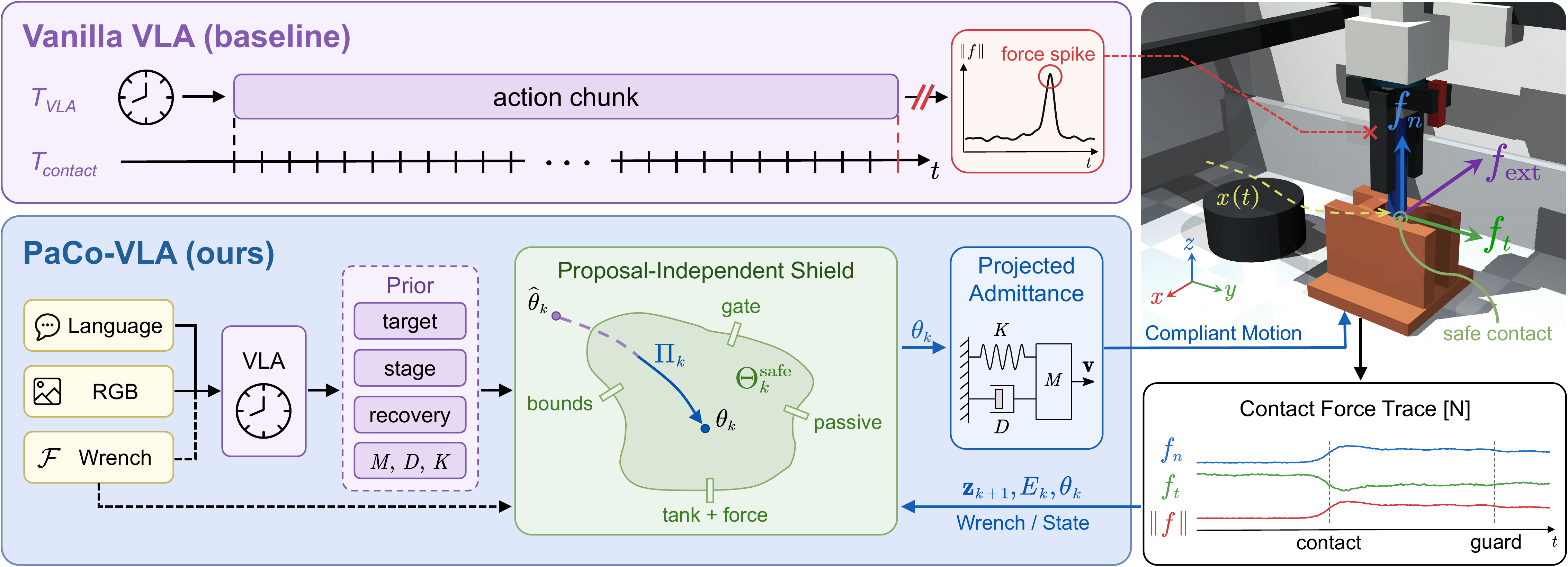}
\vspace{-2em}
\captionof{figure}{\small\textbf{PaCo-VLA overview.} Vanilla VLA sends \textit{low-rate} action chunks directly toward the plant, allowing stale predictions to persist during \textit{high-rate} contact. PaCo-VLA instead treats VLA outputs as semantic and admittance proposals, and applies only the schedule $\theta_k$ returned by a proposal-independent shield. After execution, the next contact state $\mathbf{z}_{k+1}=(\mathbf{x}_{k+1},\mathbf{v}_{k+1},\mathcal{F}_{k+1})$, tank $E_k$, and schedule $\theta_k$ seed the next projection, keeping contact authority at the shielded admittance port.}
\label{fig:paco-overview}
\end{strip}

\begin{abstract}
    Contact-rich manipulation demands both high-level semantic reasoning and the safe regulation of high-frequency contact dynamics. While Vision-Language-Action (VLA) models provide unprecedented semantic generalization, their low-rate outputs lack the reliability required for direct plant authority in force-sensitive tasks. To bridge this semantic-to-control gap, we introduce PaCo-VLA, a passivity-shielded compliance prior that recasts the VLA interface. Rather than trusting VLAs with direct motor commands, PaCo-VLA treats network outputs as task-level compliance proposals: semantic bindings, task stages, and admittance schedules. A high-frequency, proposal-independent passivity shield governs these proposals through energy-tank accounting and boundary checks, preventing invalid, stale, or unverified model predictions from bypassing low-level contact physics. This decoupled architecture also enables causal evaluation, isolating semantic contributions from geometric shortcuts. Extensive simulated and real-world experiments demonstrate that PaCo-VLA achieves superior precision over unshielded VLA baselines, sustaining zero passivity violations even under adversarial compliance shifts. This framework establishes a sampled-passive runtime contract at the admittance port and provides a runtime interface for deploying foundation models in contact-rich domains.
\end{abstract}
\vspace{-1em}

\vspace{-1em}
\section{Introduction}
	
    In recent years, embodied language and Vision-Language-Action (VLA) models have improved robot generalization by scaling language grounding, robot data, visual-language pretraining, and multi-task action learning~\cite{ahn2022saycan,brohan2022rt1, driess2023palme,zitkovich2023rt2,octo2024,kim2024openvla}. These advances make semantic transfer increasingly practical, but they leave a separate interface question unresolved: when a model emits low-rate action tokens, action chunks, or task-conditioned control proposals for a particular robot, what feedback-control contract turns those outputs into safe commands at contact time? Contact-rich manipulation makes this question acute. A robot may need language and visual context to select an object, infer a material-dependent force limit, or recover from a visual perturbation, yet the contact controller must react to wrench measurements, actuator limits, and unstable contact at a much faster time scale than a large VLA service can typically provide. Thus, we propose PaCo-VLA, a passivity-shielded compliance prior for using VLA semantics without treating direct model actions as a trustworthy plant interface. 

    The central issue is the semantics-to-contact interface rather than the expressiveness of learned actions alone. Direct action policies such as Diffusion Policy and ACT provide strong imitation-learning baselines for visuomotor control~\cite{chi2023diffusion,zhao2023act}, while open generalist policies such as Octo, OpenVLA and $\pi_{0.5}$ make language-conditioned robot actions increasingly accessible~\cite{octo2024,kim2024openvla,pi0.5}. In force-sensitive contact, a predicted action is ultimately evaluated at the contact surface, where latency, action-chunk hold time, actuator limits, force thresholds, and local compliance determine whether an otherwise plausible command remains appropriate. A delayed prediction can therefore change the effective compliance at the wrong moment, a geometrically plausible action can still exceed a contact governor, and task-level recovery can make aggregate success difficult to attribute to the semantic model, the low-level controller, or the benchmark structure. These couplings expose an interface mismatch: semantic reasoning and contact assurance operate at different rates, abstractions, and failure modes.

    We address this mismatch by separating semantic proposal generation from low-level compliance execution. Building on interaction-control and runtime assurance principles, we let the VLA propose quantities that are meaningful at the task and contact level, including semantic binding, task stage, recovery cue, and diagonal admittance schedule while the robot executes only the command produced by a proposal-independent passivity shield. The shield projects proposed parameters into admissible boxes, enforces a sampled diagonal-admittance passivity margin, charges active mass and stiffness increases to an energy tank, monitors force and torque, and enters guarded recovery or stops execution when the proposal, scene context, or runtime measurements are unsafe. Since learned, random, delayed-context, classical, and recovery proposals all pass through the same shield, the applied-command contract becomes independent of proposal source, which lets the experiments measure runtime validity and semantic contribution as separate properties rather than a single aggregate success number. Fig.~\ref{fig:paco-overview} summarizes this interface-level separation.

    This separation creates an attribution requirement for evaluation. In a shielded VLA system, high task success may reflect useful semantic information from the model, effective recovery behavior from the controller, or an unintended geometric shortcut in the benchmark. We therefore evaluate semantic value with paired simulated counterfactual trials instead of aggregate success alone. The primary condition receives live language and visual information without privileged scene labels. The control conditions then remove or corrupt the causal channels using random proposals, delayed context, shuffled language, masked images, wrong-object instructions, contradictory language, matched geometry with changed instructions, and matched instructions with changed visual targets.
Our contributions are summarized as follows:
\begin{enumerate}
    \item We introduce a VLA-to-compliance interface in which learned models propose semantic bindings and compliance schedules, while the robot executes only shielded plant commands.
    \item We develop a proposal-independent sampled admittance runtime contract combining box/passivity projections, energy-tank accounting, force monitoring, timing checks, and guarded recovery.
    \item We provide an evaluation protocol that combines runtime contract ablations, paired semantic counterfactuals, and direct-action interface tests to separate causal semantic value from shield/recovery effects and aggregate success.
\end{enumerate}

\section{Related Work}

\paragraph{Robot interaction control}
Impedance and admittance control remain foundational for contact-rich robot manipulation~\cite{hogan1985impedance}. Passivity-based methods add a system-level view of energy exchange, especially in haptic and teleoperation systems where time-domain passivity observers and controllers dissipate excess energy online~\cite{hannaford2002passivity}. Energy tanks provide a related mechanism for accounting for the active energy required by changing controller parameters~\cite{califano2022energytanks}. Our method follows this lineage but uses passivity and tank accounting as an interface around VLA-style compliance proposals~\cite{hogan1985impedance,hannaford2002passivity,califano2022energytanks}. The distinction is not that passivity alone solves contact safety, but that the proposal-independent shield sends every learned, delayed-context, random, or recovery proposal through same applied-command contract.

\paragraph{Runtime assurance and safety filters}
Runtime assurance methods often wrap an untrusted controller with a safety mechanism that prevents or modifies unsafe choices~\cite{alshiekh2018shielding}. Shielding for reinforcement learning restricts actions before they reach the system~\cite{alshiekh2018shielding}, predictive safety filters modify proposed continuous inputs using constrained model-predictive reasoning~\cite{wabersich2021predictive}, and control-barrier-function QPs encode forward-invariance conditions as real-time optimization constraints~\cite{ames2017cbf}. Our interface is closest in spirit to these safety filtering interfaces, but the contract is specialized to contact-rich VLA execution: filtered object is a semantic and compliance proposal, and the enforced invariants are parameter boxes, sampled diagonal-admittance passivity, energy-tank accounting, force monitoring, and guarded recovery.

\paragraph{Learned robot policies and VLAs}
Diffusion Policy introduced action diffusion and receding-horizon visual control as a strong visuomotor policy class~\cite{chi2023diffusion}. ACT learns action chunks for fine-grained manipulation from demonstrations~\cite{zhao2023act}. Generalist robot policies and VLAs, including Octo, OpenVLA and $\pi_{0.5}$, broaden this interface by conditioning action generation on language and large-scale multi-task data~\cite{octo2024,kim2024openvla,pi0.5}. These models motivate our proposal interface, but contact-rich execution adds feedback and force constraints that are not captured by action prediction alone~\cite{octo2024,kim2024openvla,chi2023diffusion,zhao2023act,pi0.5}. We therefore ask what runtime contract lets such models serve as useful proposal sources without making them final plant controllers.

\paragraph{Causal evaluation under shielding}
Counterfactual evaluation is a standard way to distinguish causal dependence from correlation when multiple mechanisms can explain an outcome~\cite{pearl2009causality}. Input perturbation methods similarly test which observed channels affect a model's prediction, although they do not by themselves establish robot-level causality~\cite{ribeiro2016why}. Benchmark success can also be inflated by shortcut features that solve the dataset without solving the intended semantic problem~\cite{geirhos2020shortcut}. In a shielded VLA system, these concerns combine: success may come from the learned proposal, the shield, recovery behavior, or a hidden task prior. We therefore evaluate semantic value through counterfactual controls that remove visual information, shuffle or contradict language, force delayed context, randomize proposals, or preserve geometry while changing the semantic instruction.

\section{Background}

Compliance control provides the contact-time abstraction used by PaCo-VLA. In Cartesian admittance control, the measured interaction wrench $\wrenchk=[F_k^\top,\tau_k^\top]^\top\in\mathbb{R}^6$ drives a virtual mass-spring-damper system,
\begin{equation}
    M_k\dot{\mathbf{v}}_k + D_k\vk + K_k\xk = \wrenchk ,
    \label{eq:admittance-background}
\end{equation}
where $\xk$ and $\vk$ are displacement and velocity relative to the current contact reference, and $\theta_k=(M_k,D_k,K_k)$ is the compliance schedule held over a control sample. This interface is useful because a task-level system can shape contact behavior by changing $\theta_k$ rather than commanding direct motor targets. It is also risky: rapid increases in inertia or stiffness can inject active energy into the contact loop, and passivity constrains energy exchange rather than peak instantaneous force~\cite{hogan1985impedance,hannaford2002passivity,califano2022energytanks}.

This distinction motivates treating VLA outputs as compliance priors rather than plant commands. Low-rate VLA predictions can identify the target, task stage, or material-conditioned compliance regime, but the high-rate controller must still decide which schedule is admissible at the current contact state. We therefore let the VLA propose a compliance prior $\hat{u}_k$ containing $\hat{\theta}_k$ and semantic context, while a runtime map first gates it into a candidate $\tilde{u}_k$ and then returns
\[
    \theta_k=\Pi_k(\tilde{u}_k;\theta_{k-1},E_{k-1},\zk),
    \qquad
    \zk=(\xk,\vk,\wrenchk),
\]
which produces the only schedule sent to the admittance port in \eqref{eq:admittance-background}. The rest of the method specifies this map: how the prior is represented, how it is projected into a passive sampled-admittance set, and how delayed or unsafe context is handled.

\section{Method}
\label{sec:method}

PaCo-VLA separates semantic authority from contact authority. The VLA supplies a low-rate compliance prior $\hat{u}_k$ that can encode task semantics, recovery intent, and a desired diagonal admittance schedule. A proposal-independent shield then evaluates the same candidate against the current state $\zk$, previous schedule $\theta_{k-1}$, and tank state $E_{k-1}$. Only the shielded schedule $\theta_k$ is executed, so learned, classical, delayed-context, random, and recovery proposals all face the same applied-command contract. Figure~\ref{fig:paco-gate}, Algorithms~\ref{alg:paco-project} and~\ref{alg:paco-tank} summarize the three runtime stages: semantic gating, compliance projection, and tank-guarded execution.

\subsection{Compliance Prior Interface}

\begin{figure}[t]
    \centering
    \includegraphics[width=0.95\linewidth]{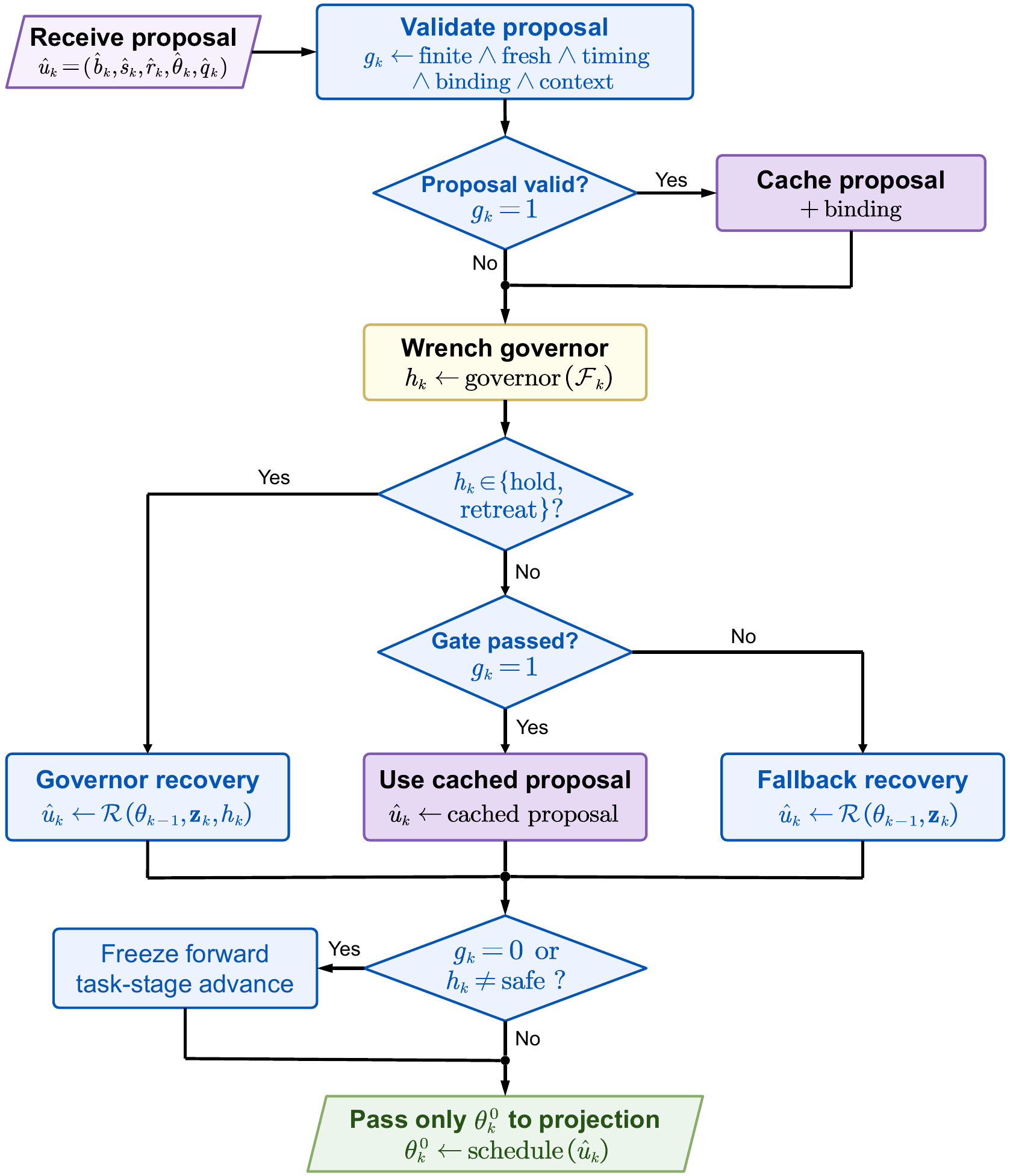}
    \vspace{-0.3em}
    \caption{\textbf{Semantic proposal gating.} Valid proposals are cached, then the wrench governor and validity gate select the proposal or recovery candidate passed to compliance projection in Algorithm~\ref{alg:paco-project}.    
    }
    \vspace{-1em}
    \label{fig:paco-gate}
\end{figure}

At each control sample, the proposal source emits
\[
    \hat{u}_k =
    \left(\hat{b}_k,\hat{s}_k,\hat{r}_k,\hat{\theta}_k,\hat{q}_k\right),
\]
where \(\hat{b}_k\) is the semantic binding, \(\hat{s}_k\) is the task stage, \(\hat{r}_k\) is a recovery cue, \(\hat{\theta}_k=(\hat{M}_k,\hat{D}_k,\hat{K}_k)\) is the proposed diagonal admittance schedule, and \(\hat{q}_k\) records validity quantities such as freshness, context consistency, and residual estimates. For the learned proposal source, the frozen VLA backbone encodes the synchronized RGB-language observation into a language-visual representation, from which the trainable fusion module jointly decodes \(\hat{b}_k,\hat{s}_k,\hat{r}_k,\hat{\theta}_k,\) and \(\hat{q}_k\). The semantic binding \(\hat{b}_k\) specifies the task-relevant object, slot, and material references. The stage output selects a shared motion primitive and contributes a bounded prior term. The learned proposal is carried with the corresponding binding and used by the validity gate for consistency checking.
The proposal gate accepts finite, fresh proposals whose context remains consistent with the current task; otherwise it supplies a conservative recovery candidate, as summarized in Figure~\ref{fig:paco-gate}. This representation assigns the VLA to semantic selection and task-level compliance choice, while preventing unfiltered network outputs from becoming motor commands.

\begin{figure*}[t]
    \centering
    \includegraphics[width=0.95\linewidth]{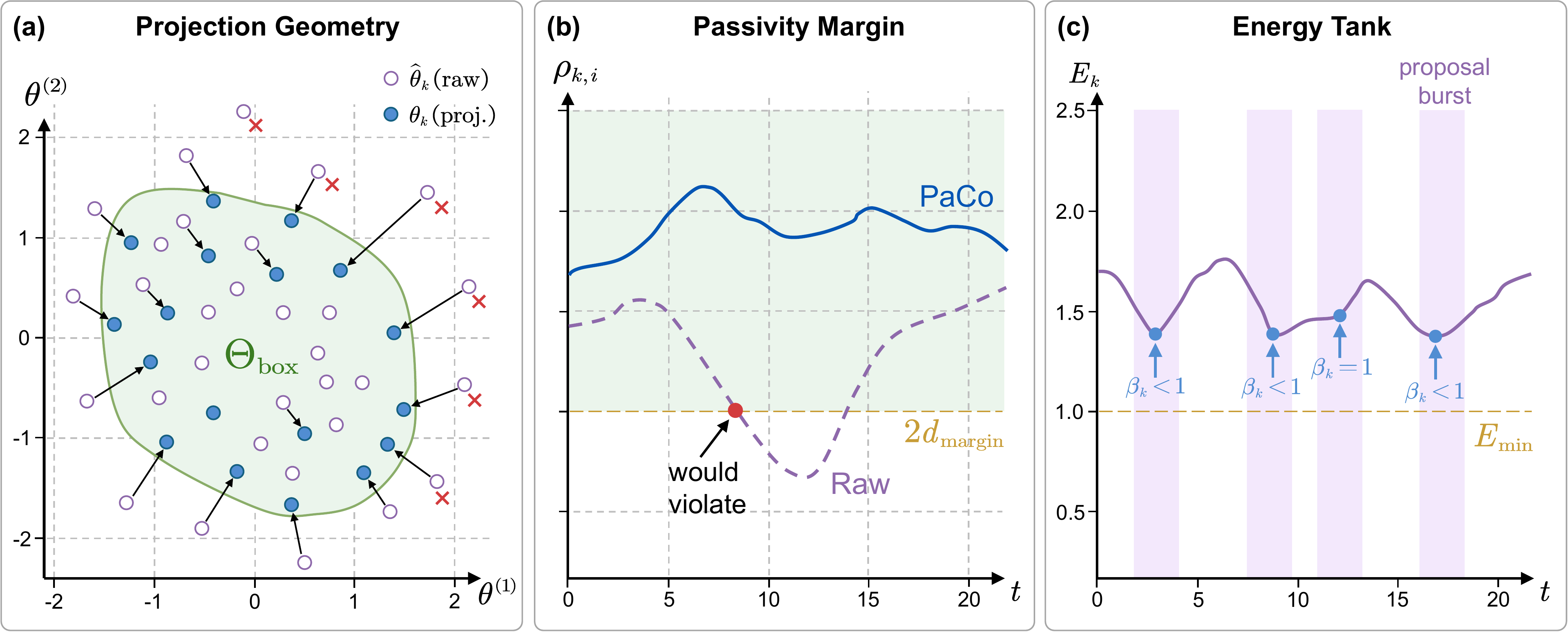}
    \vspace{-0.6em}
    \caption{\textbf{Runtime shield mechanisms.}
    (a) Box projection maps unfiltered proposals into $\Theta_{\mathrm{box}}$; (b) margin projection enforces $\rho_{k,i}(\theta_k^{\mathrm{pass}};\theta_{k-1})\ge2d_{\mathrm{margin}}$; (c) the tank applies $\beta_k\in[0,1]$. Here $\beta_k=1$ accepts the projected schedule, while $\beta_k<1$ scales active changes to keep $E_k\ge E_{\min}$.}
    \vspace{-0.6em}
    \label{fig:shield-mechanisms}
\end{figure*}

\subsection{Proposal-Independent Shield}
The shield converts any finite candidate schedule into an executable schedule through a fixed sequence of checks. Let $\Theta_{\mathrm{box}}$ denote the configured admissible intervals, with positive mass and damping and nonnegative stiffness. Given $\theta^0_k$ from Figure~\ref{fig:paco-gate}, Algorithm~\ref{alg:paco-project} first clips mass, damping, and stiffness into $\Theta_{\mathrm{box}}$ to obtain $\theta_k^{\mathrm{box}}$. Its margin stage then returns $\theta_k^{\mathrm{pass}}$ by enforcing the sampled diagonal-admittance margin
\begin{equation}
    \begin{aligned}
    \rho_{k,i}(\theta;\theta_{k-1})
    &= 2D_i(\theta) - \frac{M_i(\theta)-M_{k-1,i}}{\Delta t_k} \\
    &\ge 2d_{\mathrm{margin}},\qquad i=1,\ldots,6.
    \end{aligned}
    \label{eq:passivity-margin}
\end{equation}
The projection prefers to raise damping; if the damping box cannot support the requested positive mass jump, it reduces that jump. This asymmetric rule is intentional: decreasing mass is not charged as active inertial storage, while rapidly increasing mass must be backed by damping at same sample.

\begin{algorithm}[htbp]
\caption{Compliance projection}
\label{alg:paco-project}
\begin{algorithmic}[1]

\State $\theta_k^{\mathrm{box}}\gets\mathrm{clip}(\theta^0_k,\Theta_{\mathrm{box}})$
\State $\theta_k^{\mathrm{pass}}\gets\theta_k^{\mathrm{box}}$

\For{axis $i=1,\ldots,6$}
    \State $(m_i,d_i)\gets(M_{k,i}^{\mathrm{box}},D_{k,i}^{\mathrm{box}})$
    \State $\rho_i\gets2d_i-(m_i-M_{k-1,i})/\Delta t_k$
    \If{$\rho_i<2d_{\mathrm{margin}}$}
        \State raise $d_i$ within damping box; update $\rho_i$
    \EndIf
    \If{$\rho_i<2d_{\mathrm{margin}}$}
        \State reduce $[m_i-M_{k-1,i}]_+$; update $\rho_i$
    \EndIf
    \State $(M_{k,i}^{\mathrm{pass}},D_{k,i}^{\mathrm{pass}})\gets(m_i,d_i)$
\EndFor

\State keep $K_k^{\mathrm{pass}}$ inside the stiffness box
\State \Return $\theta_k^{\mathrm{pass}}$ satisfying boxes and margin
\end{algorithmic}
\end{algorithm}

\begin{algorithm}[htbp]
\caption{Tank guarded execution}
\label{alg:paco-tank}
\begin{algorithmic}[1]

\AlgStage{Stage 1: Credit passive dissipation}
\State $P_{D,k}\gets\Delta t_k\vk^\top D_{k-1}\vk$
\State $E_k^-\gets\min(E_{\max},E_{k-1}+P_{D,k})$
\vspace{-0.4em}
\AlgStage{Stage 2: Compute active-storage charge}
\State $\Delta M_i^+\gets[M_{k,i}^{\mathrm{pass}}-M_{k-1,i}]_+$
\State $\Delta K_i^+\gets[K_{k,i}^{\mathrm{pass}}-K_{k-1,i}]_+$
\State $A_k\gets \frac{1}{2}\sum_i\Delta M_i^+\bar{v}_{k,i}^2 + \frac{1}{2}\sum_i\Delta K_i^+\bar{x}_{k,i}^2$

\vspace{-0.4em}
\AlgStage{Stage 3: Compute tank-feasible scale}
\If{$A_k=0$}
    \State $\beta_k\gets1$
\Else
    \State $\beta_k\gets\min\!\left(1,\frac{[E_k^- - E_{\min}]_+}{\alpha A_k}\right)$
\EndIf

\vspace{-0.4em}
\AlgStage{Stage 4: Apply guarded schedule}
\State $\theta_k\gets\theta_{k-1}+\beta_k(\theta_k^{\mathrm{pass}}-\theta_{k-1})$
\State $E_k\gets E_k^- - \alpha\beta_kA_k$
\State execute admittance control; block raw VLA output
\end{algorithmic}
\end{algorithm}

The energy tank then limits how much of the remaining active parameter change can be applied. It first credits conservative damping dissipation from the previous applied schedule,
\[
    \begin{aligned}
    P_{D,k}&=\Delta t_k\vk^\top D_{k-1}\vk, \\
    E_k^-&=\min(E_{\max},E_{k-1}+P_{D,k}).
    \end{aligned}
\]
With per-axis scalar envelopes $\bar{x}_{k,i}$ and $\bar{v}_{k,i}$, the active storage charge is
\begin{equation}
    \begin{aligned}
    A_k(\theta,\theta_{k-1})
    &= \frac{1}{2}\sum_i [M_{k,i}-M_{k-1,i}]_+\bar{v}_{k,i}^2 \\
    &\quad + \frac{1}{2}\sum_i [K_{k,i}-K_{k-1,i}]_+\bar{x}_{k,i}^2 .
    \end{aligned}
    \label{eq:active-storage}
\end{equation}
where $[a]_+=\max(a,0)$. The charge in \eqref{eq:active-storage} counts only positive mass and stiffness changes, matching the active storage that can be injected by a parameter switch. Let $\theta_k^{\mathrm{pass}}$ denote the schedule after box and margin projection, and abbreviate $A_k=A_k(\theta_k^{\mathrm{pass}},\theta_{k-1})$ below. Algorithm~\ref{alg:paco-tank} applies the largest tank-feasible interpolation,
\begin{equation*}
    \begin{aligned}
    \theta_k &=\theta_{k-1}+\beta_k(\theta_k^{\mathrm{pass}}-\theta_{k-1}), \\
    \beta_k &=
    \begin{cases}
        1, & A_k=0,\\
        \min\!\left(1,\dfrac{[E_k^- - E_{\min}]_+}{\alpha A_k}\right), & A_k>0,
    \end{cases}
    \end{aligned}
\end{equation*}
with $\beta_k\in[0,1]$, $\alpha\ge1$, and updated tank state
$E_k=E_k^- - \alpha\beta_kA_k(\theta_k^{\mathrm{pass}},\theta_{k-1})$.
Fig.~\ref{fig:shield-mechanisms} illustrates the geometry of projection,
margin enforcement, and tank scaling. 
Appendix\ref{app:runtime-contract-sketch}
states the corresponding sampled storage contract and proof.

\subsection{Guarded Execution}

Guarded execution addresses failure modes outside the projection operator: missing proposals, delayed semantic context, inconsistent visual binding, and measured wrench excursions. Let $g_k\in\{0,1\}$ denote the proposal gate after finiteness, freshness, and context checks, and let $h_k$ denote the wrench-governor state. Write $\mathcal{H}$ for intervention states, including hold, retreat, and safety-stop decisions. The candidate sent to the shield is
\[
    \begin{aligned}
    \tilde{u}_k &=
    \begin{cases}
        \mathcal{R}(\theta_{k-1},\zk,h_k), & h_k\in\mathcal{H},\\
        \hat{u}_k, & h_k\notin\mathcal{H}\ \text{and}\ g_k=1,\\
        \mathcal{R}(\theta_{k-1},\zk), & h_k\notin\mathcal{H}\ \text{and}\ g_k=0,
    \end{cases} \\
    \theta_k&=\Pi_k(\tilde{u}_k;\theta_{k-1},E_{k-1},\zk).
    \end{aligned}
\]
The recovery map $\mathcal{R}$ keeps or reduces active inertia, raises damping within the admissible box, lowers stiffness toward a safe-hold value, and pauses task-stage advancement until a valid context returns. A calibrated force/torque governor monitors $\wrenchk$ in parallel; upon activation, it issues velocity scaling and recovery or safety-stop decisions before any new task-stage transition. Thus delayed or unsafe semantic outputs remain observable proposal failures and do not directly command the plant.

\medskip
\noindent\textbf{Runtime contract.}
Together, the projection, tank update, and gate define the applied-command contract: every finite proposal is transformed into a bounded sampled-admittance schedule and applied only through the admittance port. 
Appendix\ref{app:runtime-contract-sketch} states the sampled storage claim and proof.

\section{Experiments}

\subsection{Experimental Setup}

\paragraph{Tasks and metrics}
We evaluate PaCo-VLA on contact-rich tasks in simulation and on a real robot. The experiments cover a sampled runtime-contract check, classical-control comparisons, counterfactual semantic attribution, direct learned-action comparisons, a cross-task simulated wiping study and a real-robot connector-insertion study. 
Task success records completion of the configured insertion depth and lateral tolerance for connector insertion, and selecting the target and wiping profile specified by the instruction for wiping.
Safety rate records the fraction of trials that remain within the configured force/torque and sampled-passivity limits throughout execution, independent of task completion.

\paragraph{Hardware platform}
We use an AUBO-i5 arm with a DH AG-160-95 gripper, a wrist Intel D435i camera, an external D415 camera, and a KWR75B six-axis force/torque sensor. The robot executes real-world tasks under the same guarded-compliance interface used in simulation evaluation.

\paragraph{Proposal generator} We instantiate the learned proposal generator with a frozen \textit{OpenVLA-7B} backbone and a 2.57M parameter fusion head that jointly decodes the structured proposal \(\hat{u}_k\) from the backbone language-visual representation. The stage-prior blending weight is capped at 0.35. More training and input details, including hyperparameters and the prompt, are provided in Appendix\ref{app:proposal-generator}.

\paragraph{Baselines}
We compare against fixed and adaptive admittance controllers, a passivity observer, an energy-tank, a force-scheduling, a nonsemantic shielded controller, and direct learned-action policies including OpenVLA, Octo, Diffusion Policy, ACT, and $\pi_{0.5}$. Learned policies use a common 30-episode task-adaptation set and episode-level train/validation split, using native architectures/objectives.

\subsection{Runtime Contract and Classical Controls}

Table~\ref{tab:runtime} reports the runtime contract check and the classical control comparison. The contract check samples adversarial compliance proposals, and the controller comparison uses matched simulated connector trials. The complete projection has \textbf{0/1000} sampled passivity-margin violations under adversarial proposals; removing passivity projection violates the margin in \textbf{50.5\%} of trials. In simulated contact, \methodname{} has the highest task success among the classical controllers.

\begin{table}[htbp]
\centering
\caption{\textbf{Runtime projection and classical-control results.} }
\vspace{-1em}
\label{tab:runtime}
\footnotesize

\begin{tabular}{@{}l@{\hspace{1em}}rrr@{}}
\toprule
\textbf{Condition} & \textbf{Task $\uparrow$} &
\textbf{Safe $\uparrow$} & \textbf{Viol. $\downarrow$} \\
\midrule
\multicolumn{4}{@{}l}{\emph{Sampled contract, N=1000 each}} \\
No Projection & -- & -- & \bad{0.505} \\
\rowcolor{oursrow}
\methodname{} + Projection & -- & -- & \best{0.000} \\
\midrule
\multicolumn{4}{@{}l}{\emph{Controller comparison, N=192 each}} \\
Fixed admittance~\cite{hogan1985impedance} & 0.000 & 0.979 & 0.000 \\
Rule admittance~\cite{hogan1985impedance} & 0.255 & \best{1.000} & 0.000 \\
Passivity observer~\cite{hannaford2002passivity} & 0.250 & 0.990 & 0.000 \\
Energy tank~\cite{califano2022energytanks} & \underline{0.260} & 0.995 & 0.000 \\
Force scheduling~\cite{raibert1981hybrid} & 0.000 & 0.958 & 0.000 \\
\rowcolor{oursrow}
\methodname{} proposals & \best{0.318} & 0.953 & \textbf{0.000} \\
\bottomrule
\end{tabular}

\par\vspace{2pt}
\parbox{\linewidth}{%
    \footnotesize\raggedright \qquad \qquad
    \textit{Note. } Viol. denotes sampled passivity violations.\par}
\end{table}

\vspace{-1.5em}

\begin{table}[htbp]
\centering
\caption{\textbf{Counterfactual semantic evaluation.} }
\vspace{-1em}
\label{tab:semantic}
\footnotesize

\begin{tabular}{@{}l@{\hspace{1em}}rrr@{}}
\toprule
\textbf{Mode} & \textbf{Task $\uparrow$} &
\textbf{Safe $\uparrow$} & \textbf{Valid $\uparrow$} \\
\midrule
\multicolumn{4}{@{}l}{\emph{Semantic channels available, N=144 each}} \\
Oracle & 0.993 & 0.993 & 0.993 \\
\rowcolor{oursrow}
\textbf{Live L+V} & \best{0.979} & \textbf{1.000} & \textbf{1.000} \\
\midrule
\multicolumn{4}{@{}l}{\emph{Counterfactual semantic controls, N=144 each}} \\
Force-only baseline~\cite{raibert1981hybrid} & 0.000 & 1.000 & 0.597 \\
Shuffled L & 0.000 & 1.000 & 1.000 \\
Masked image & 0.000 & 1.000 & 1.000 \\
Wrong object & 0.000 & 1.000 & 1.000 \\
Contradictory L & 0.000 & 1.000 & 1.000 \\
Same geom., diff. instr. & 0.000 & 1.000 & 1.000 \\
Same instr., diff. target & 0.000 & 1.000 & 1.000 \\
Stale context & 0.000 & 0.986 & 0.986 \\
Random proposal & 0.000 & 0.792 & 0.792 \\
\bottomrule
\end{tabular}

\par\vspace{2pt}
\parbox{\linewidth}{%
    \footnotesize\raggedright \qquad \qquad
    \textit{Note. }Live denotes online synchronized inputs;
    L denotes \\ \qquad \qquad language and V denotes vision.\par}
\end{table}

\vspace{-1em}
\subsection{Counterfactual Semantic Attribution}
\vspace{-1em}
\begin{figure}[htbp]
    \centering
    \includegraphics[width=0.45\textwidth]{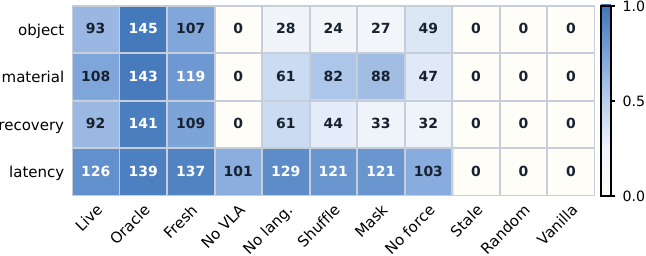}
    \vspace{-1.2em}
    \caption{\textbf{Semantic attribution.}}
    \vspace{-1em}
    \label{fig:semantic-heatmap}
\end{figure}

\begin{table*}[t]
\centering
\caption{\textbf{Direct learned-action comparison across policy backbones.}}
\vspace{-1em}
\label{tab:learned}

\small
\setlength{\tabcolsep}{4pt}
\renewcommand{\arraystretch}{1.08}

\begin{tabular}{
@{}l
cc c
@{\hspace{0.9em}}
cc cc cc cc c
@{}
}
\toprule

&
\multicolumn{3}{c}{
  \makebox[0pt][c]{
    \textbf{Action-token interface}
  }
}
&
\multicolumn{9}{c}{
  \makebox[0pt][c]{
    \textbf{Direct learned-action matrix}
  }
}
\\

\cmidrule(lr){2-4}
\cmidrule(lr){5-13}
\morecmidrules
\cmidrule(lr){2-4}
\cmidrule(lr){5-13}

&
\multicolumn{2}{c}{
  \textbf{OpenVLA}~\cite{kim2024openvla}
}
&
\textbf{\methodname{}}
&
\multicolumn{2}{c}{
  \textbf{Octo}~\cite{octo2024}
}
&
\multicolumn{2}{c}{
  \textbf{DP}~\cite{chi2023diffusion}
}
&
\multicolumn{2}{c}{
  \textbf{ACT}~\cite{zhao2023act}
}
&
\multicolumn{2}{c}{
  \boldmath$\boldsymbol{\pi}_{0.5}$~\cite{pi0.5}
}
&
\textbf{\methodname{}}
\\

\cmidrule(lr){2-3}
\cmidrule(lr){4-4}
\cmidrule(lr){5-6}
\cmidrule(lr){7-8}
\cmidrule(lr){9-10}
\cmidrule(lr){11-12}
\cmidrule(lr){13-13}

&
\textbf{w/o} & \textbf{w/}
& \textbf{Ours}
&
\textbf{w/o} & \textbf{w/}
&
\textbf{w/o} & \textbf{w/}
&
\textbf{w/o} & \textbf{w/}
&
\textbf{w/o} & \textbf{w/}
& \textbf{Ours}
\\

\midrule

\textbf{Task} $\uparrow$
&
0.0000 & 0.0075
& \cellcolor{oursrow}\best{0.2408}
&
0.0000 & 0.0000
&
0.0139 & 0.0139
&
0.0694 & 0.0833
&
0.2639 & 0.2639
& \cellcolor{oursrow}\best{0.7361}
\\

\textbf{Safe} $\uparrow$
&
0.7308 & 0.8267
& \cellcolor{oursrow}\best{1.0000}
&
1.000 & 1.000
&
\underline{0.993} & \underline{1.000}
&
1.000 & 1.000
&
\underline{0.875} & \underline{0.958}
& \cellcolor{oursrow}\best{1.000}
\\

\bottomrule
\end{tabular}
\par\vspace{2pt}
\parbox{\linewidth}{%
    \footnotesize \qquad \quad
    \textit{Note. }Token Interface: $N=1200$; Direct Action: $N=144$. ``w/o'' and ``w/'' denote without and with safety projection, respectively. \par}
\end{table*}

 The counterfactual semantic study spans 1584 paired conditions across object binding, material-conditioned compliance, and visual recovery. Table~\ref{tab:semantic} attributes success to live VLA-derived semantic information rather than recovery shortcuts. The live language-and-visual condition succeeds in \textbf{141/144} trials, close to oracle scheduling; every counterfactual semantic control has zero task success under matched task instances. Fig.~\ref{fig:semantic-heatmap} gives the corresponding stratified view across semantic families and counterfactual modes. All modes use the same shielded admittance interface and plant.

 \begin{figure*}[htbp]
    \centering
    \vspace{-0.5em}
    \includegraphics[width=\linewidth]{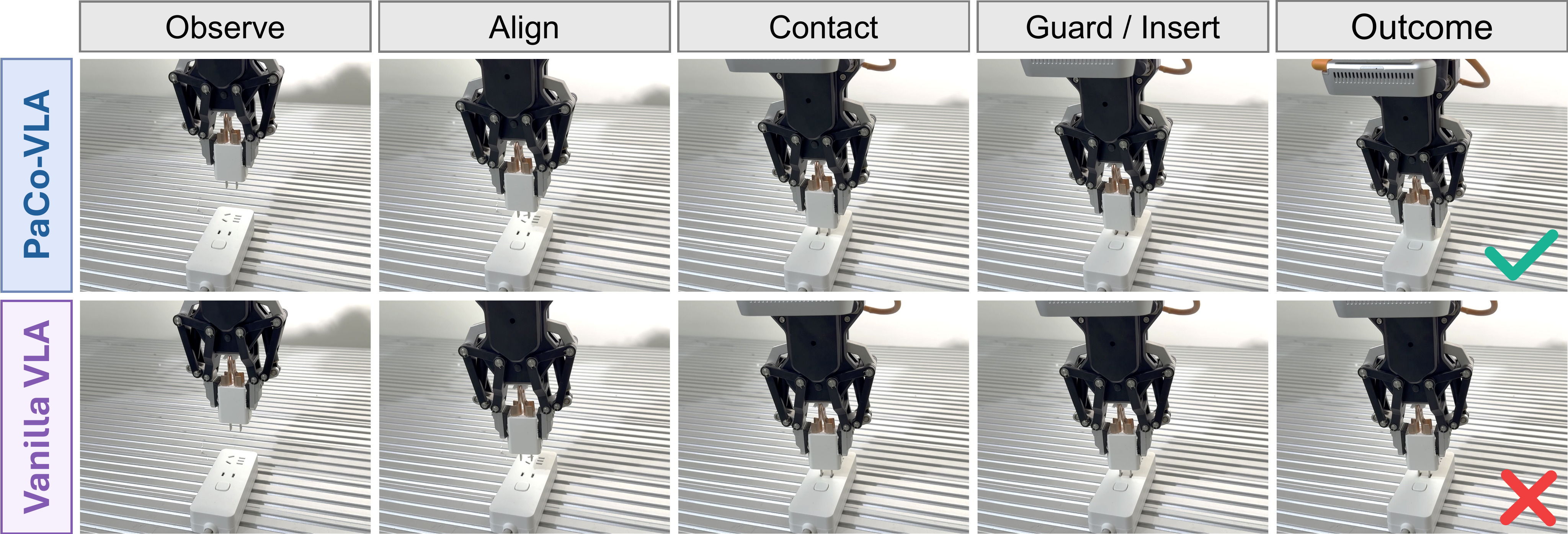}
    \vspace{-2em}
    \caption{\textbf{Real-robot contact execution sequence.}
    Stage-aligned visual evidence from representative \tracepaco{PaCo-VLA} success and
    \tracevanilla{vanilla VLA} failure trials.}
    \label{fig:real_robot_execution_sequence}
    \vspace{-1.3em}
\end{figure*}

\subsection{Direct Learned-Action Comparison}

Table~\ref{tab:learned} first compares the native OpenVLA action-token interface with PaCo-VLA using the same backbone and adaptation data, directly contrasting the policy-to-plant execution interface. The broader matrix extends this comparison to task-adapted Octo, Diffusion Policy, ACT, and $\pi_{0.5}$ under the stated contact budget and common success and safety metrics. Direct policies issue actions through their native interfaces, whereas PaCo-VLA exposes shielded semantic and admittance proposals.

\subsection{Cross-task Simulation}
On another separate simulated wiping task with fresh RGB/language inputs, PaCo-VLA reaches \textbf{140/144} instructed target/profile successes versus 9/144 for a matched $\theta$-only controller, while both controllers complete 144/144 nominal paths. This provides a cross-task check of semantic target/profile selection under sustained tangential contact.

\subsection{Real-Robot Connector Insertion}

\paragraph{Balanced trials}
\begin{wraptable}[5]{r}{0.5\columnwidth}
    \vspace{-1.8em}
    \centering
    \caption{\textbf{Real-robot success.}}
    \label{tab:real_robot_success_brief}
    \vspace{-0.8em}

    {\footnotesize
    \setlength{\tabcolsep}{3pt}
    \begin{tabular}{@{}lcc@{}}
    \toprule
    Method & Contact & Insert \\
    \midrule
    Nonsemantic shield & 16/30 & 11/30 \\
    Energy tank & 20/30 & 18/30 \\
    \rowcolor{oursrow}
    \methodname{} & \best{30/30} & \best{27/30} \\
    \bottomrule
    \end{tabular}
    }
\end{wraptable}

Fig.~\ref{fig:real_robot_execution_sequence} shows the stage-aligned real-robot
contact sequence, contrasting a completed \tracepaco{PaCo-VLA} insertion with an \tracevanilla{vanilla VLA} failure over the same visual phases. The 90-trial balanced connector
protocol compares \tracepaco{PaCo-VLA} with a nonsemantic shielded controller and an
energy-tank baseline under the same connector fixture and force/torque governor
thresholds. Table~\ref{tab:real_robot_success_brief} summarizes contact and
insertion success. \tracepaco{PaCo-VLA} also reduces final lateral error against both baselines,
reaching
$0.181\pm0.108$ mm compared with $0.622\pm0.169$ mm and $0.617\pm0.148$ mm,
with no warning or termination threshold force violations across the protocol.
\vspace{-0.5em}
\begin{figure}[htbp]

    \centering
    \includegraphics[width=\linewidth]{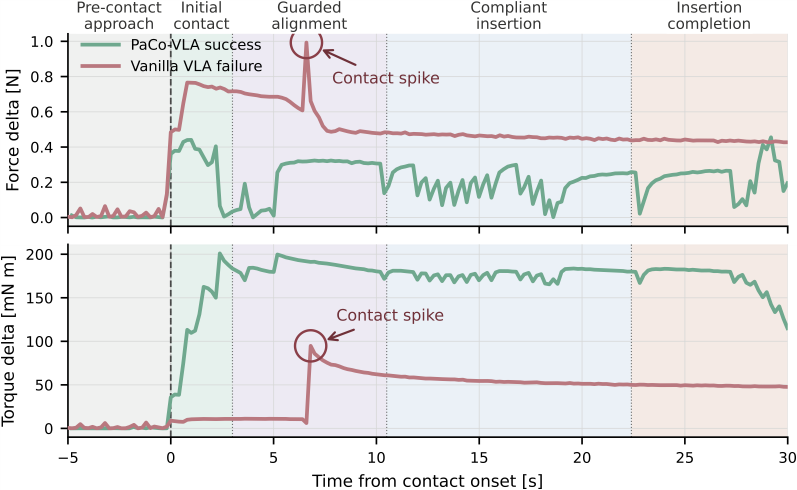}
    \vspace{-2em}
    \caption{\textbf{Contact-aligned force/torque traces.}}
    \label{fig:success-failure-contact-ft}
    \vspace{-1em}
\end{figure}

\begin{figure*}[t]
    \centering
    \includegraphics[width=\linewidth]{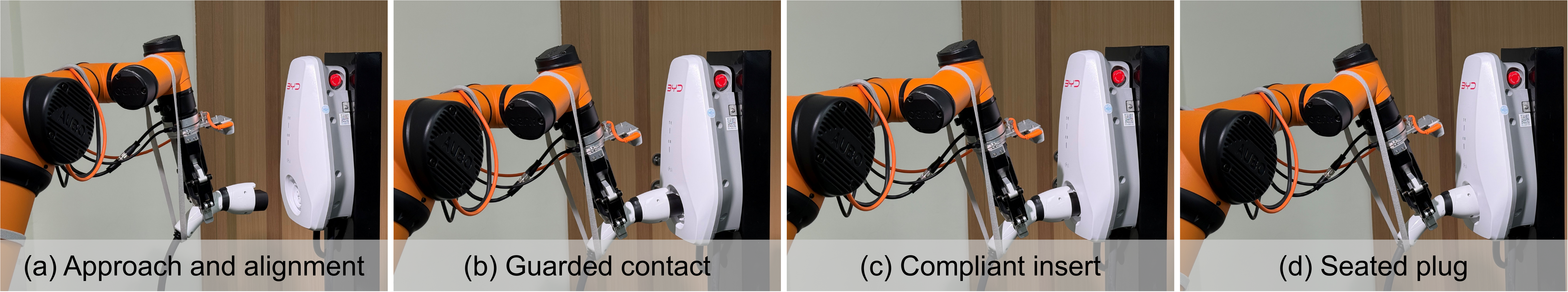}
    \vspace{-2em}
    \caption{\textbf{Real-robot EV charging-gun plugging.}
    Representative sequence for the larger EV charging-gun transfer setting.}
    \label{fig:ev_charging_sequence}
    \vspace{-1.9em}
\end{figure*}

\paragraph{Contact traces}
The contact-aligned force/torque traces in
Fig.~\ref{fig:success-failure-contact-ft} explain the corresponding contact
behavior. Both runs are aligned at initial contact, so the following phases
compare guarded alignment and insertion rather than approach timing. The failed
\tracevanilla{vanilla} run develops a sharp force/torque transient during guarded alignment
and remains on a failed contact trajectory, whereas \tracepaco{PaCo} maintains contact
through guarded alignment and compliant insertion without triggering the
force/torque governor. This links the visual outcome in
Fig.~\ref{fig:real_robot_execution_sequence} to contact evidence.

\paragraph{EV transfer}
Fig.~\ref{fig:ev_charging_sequence} shows a larger EV charging-gun plugging
sequence using the same guarded-compliance interface. 
PaCo-VLA achieves \textbf{17/30} successful EV charging-gun plug-ins, outperforming \(\pi_{0.5}\), which succeeds in 4/30 trials.

\section{Conclusion}
\label{sec:conclusion}
In this paper, we introduced PaCo-VLA, a passivity-shielded compliance interface for contact-rich
VLA manipulation. The key idea is to let the VLA provide semantic bindings, task stages, and
compliance proposals, while a proposal-independent runtime shield retains admittance authority.
Simulation, semantic counterfactuals, direct-action comparisons, and real-robot connector and EV
charging-gun trials show that PaCo-VLA preserves the sampled passivity margin, reveals when
semantics matter, and improves contact-rich execution over direct-action and baselines.

\appendices

\section*{Appendix}
\addcontentsline{toc}{section}{Appendix}

\subsection{Sampled Runtime Contract}
\label{app:runtime-contract-sketch}

We state the sampled storage guarantee underlying the runtime shield.
Assume that the admittance is diagonal and sample-and-hold over
$[t_{k-1},t_k]$, all states and proposals are finite,
$D_{\min,i}\ge d_{\mathrm{margin}}$, and
$|x_{k,i}|\le\bar{x}_{k,i}$,
$|v_{k,i}|\le\bar{v}_{k,i}$.
For the previous applied schedule, assume the fixed-parameter update satisfies

\begin{equation}
\begin{aligned}
    &V_k(M_{k-1},K_{k-1})-V_{k-1}(M_{k-1},K_{k-1}) \\
&\quad\le
\wrenchk^\top\vk\Delta t_k-P_{D,k}
+\varepsilon_k^{\mathrm{num}},
\label{A1}
\end{aligned}
\end{equation}

where
$V_k(M,K)=\frac12\vk^\top M\vk+\frac12\xk^\top K\xk$
and $0\le E_k^- -E_{k-1}\le P_{D,k}$.

\textbf{Theorem 1 (Proposal-independent sampled contract).}
After projection and tank interpolation, every nonterminated sample
remains inside the parameter boxes, satisfies
\eqref{eq:passivity-margin}, and preserves
$E_{\min}\le E_k\le E_{\max}$. Moreover, with
$S_k=V_k(M_k,K_k)+E_k$,
\begin{equation}
    S_k-S_{k-1}
\le
\wrenchk^\top\vk\Delta t_k+\varepsilon_k^{\mathrm{num}},
\label{A2}
\end{equation}
independently of the proposal source.

\emph{Proof.}
Box feasibility follows because the applied schedule is a convex
interpolation between two feasible schedules. The margin in
\eqref{eq:passivity-margin} is affine in $(M,D)$; both
$\theta_{k-1}$ and $\theta_k^{\mathrm{pass}}$ satisfy it, hence so
does their interpolation.
At the switching state, the parameter-induced storage increase obeys
\(
V_k(M_k,K_k)-V_k(M_{k-1},K_{k-1})
\le \beta_k A_k .
\)
The tank rule gives
$\alpha\beta_kA_k\le E_k^- -E_{\min}$ and therefore preserves the
tank bounds. Since $\alpha\ge1$,
\(
V_k(M_k,K_k)-V_k(M_{k-1},K_{k-1})
+E_k-E_{k-1}\le P_{D,k}.
\)
Combining this inequality with (\ref{A1}) yields
(\ref{A2}). \hfill$\square$

\subsection{Proposal Generator and Training}
\label{app:proposal-generator}

The learned proposal generator in Section~\ref{sec:method} uses a frozen
\textit{OpenVLA-7B} backbone with a trainable fusion module
containing 2.57M parameters. The input prompt is
``\texttt{In: What action should the robot take to
\{instruction\}? Out:}''.
The fusion module jointly predicts the semantic binding, task stage, recovery cue,
an 18D log-admittance vector
($\log \theta = [\log \hat{M}_{k,1:6},\log \hat{D}_{k,1:6},
 \log \hat{K}_{k,1:6}]$), and six auxiliary diagnostics:
1) residual, 2) passivity margin, 3) force risk, 4) fallback, 5) freshness, and 6) task-success confidence.
These outputs form the structured proposal \(\hat{u}_k\).

The training dataset comprises 30 episodes and 1,011 observation windows; episodes are split before window extraction, yielding a 693/132/186-window train/validation/test split.
We optimize the fusion module for 13 epochs using AdamW
with a learning rate of $10^{-3}$ and a batch size of 32.
The semantic binding, stage, recovery, admittance, and diagnostic outputs are trained jointly with a multitask objective combining cross-entropy, Huber, and binary cross-entropy losses. 

\subsection{Runtime Gate and Recovery}
\label{app:runtime-gate}
The implementation uses the parameter choices in
Table~\ref{tab:runtime-parameters}.

\begin{table}[htbp]
\vspace{-0.2em}
\caption{\textbf{Default runtime parameters used in the implementation.}}
\vspace{-1em}
\label{tab:runtime-parameters}
\centering
\footnotesize
\setlength{\tabcolsep}{2.5pt}
\renewcommand{\arraystretch}{1.08}
\begin{tabularx}{\columnwidth}{
    @{}
    >{\raggedright\arraybackslash}p{0.28\columnwidth}
    >{\raggedright\arraybackslash}p{0.18\columnwidth}
    >{\raggedright\arraybackslash}p{0.27\columnwidth}
    >{\raggedright\arraybackslash}X
    @{}
}
\toprule
\textbf{Parameter} &
\textbf{Default} &
\textbf{Parameter} &
\textbf{Default} \\
\midrule

Safety projection
& 250 Hz
& Mass box
& $M_i\in[0.1,10]$ \\

Compliance control
& 250Hz 
& Damping box
& $D_i\in[5,500]$ \\

Policy proposals
& 50 Hz
& Stiffness box
& $K_i\in[0,2000]$ \\

Camera observations
& 10 Hz
& OpenVLA service
& 0.20 s \\

Passivity margin
& $d_{\mathrm{margin}}=2$
& timeout
& ~\\

Initial tank energy
& $E_0=10$
& Stale VLA proposal
& 0.30 s \\

Tank charge factor
& $\alpha=1.2$
& Stale F/T sample
& 0.05 s \\

Tank energy bounds
& \begin{tabular}[t]{@{}l@{}}
  $E_{\min}=0.5$\\[-1pt]
  $E_{\max}=20$
  \end{tabular}
& Stale compliance command
& 0.30 s \\

\bottomrule
\end{tabularx}
\end{table}
\vspace{-0.5em}

The force/torque governor records threshold, impulse, and impact events. Its
default runtime thresholds are 15 N soft and 25 N hard lateral force, 1.0 Nm
soft and 2.0 Nm hard torque, 6.0 Ns lateral impulse, 1.2 Nms torque impulse,
and 6000 N/s impact rate; the corresponding velocity scales are 0.40 for soft
events, 0.30 for impact events, and 0 for hard-stop events. The recorded trace includes
proposal source, freshness, context consistency, semantic binding validity,
recovery indicator, residual estimate, tank scale $\beta_k$, tank state $E_k$,
projected schedule $\theta_k$, passivity-margin minimum, and force-governor
state. A latched semantic target may be used after a previously accepted fresh
proposal, but it is explicitly recorded as latched continuity rather than a new
fresh model-forward proposal. These variables support semantic attribution
because the proposal source can change while the same schedule projection and
contact-time guards remain in use.

\subsection{Simulation Protocol}
\label{app:simulation-protocol}
The simulated evaluation is organized into four groups. The runtime-contract
group samples 1000 adversarial compliance proposals per condition and measures
violations of \eqref{eq:passivity-margin}. The classical-control group uses
192 matched connector trials per controller, with the same contact thresholds,
initial-state distribution, and success criteria within that group. The semantic
group uses 11 modes, 3 semantic families, 3 difficulty levels, and 16 paired
replicates; paired trials share the same task instance while language, vision,
proposal freshness, target identity, or proposal source is changed according to
the counterfactual condition. The direct-action group evaluates task-adapted policies on matched task instances through their native action interfaces under the common contact and safety protocol.

\subsection{Compute Resources}
\label{app:compute-resources}
Neural VLA inference and learned-policy evaluation used 2 $\times$ NVIDIA GeForce RTX 4090
GPUs, with OpenVLA evaluated in bfloat16. The 250 Hz passivity shield,
classical controllers, force/torque logging, offline statistics, and real-robot
servo loop were run without GPU acceleration. RTX 4090 GPUs were used only for
neural-model evaluation, not for the low-level runtime contract.

\begin{figure}[htbp]
    \centering
    \includegraphics[width=\linewidth]{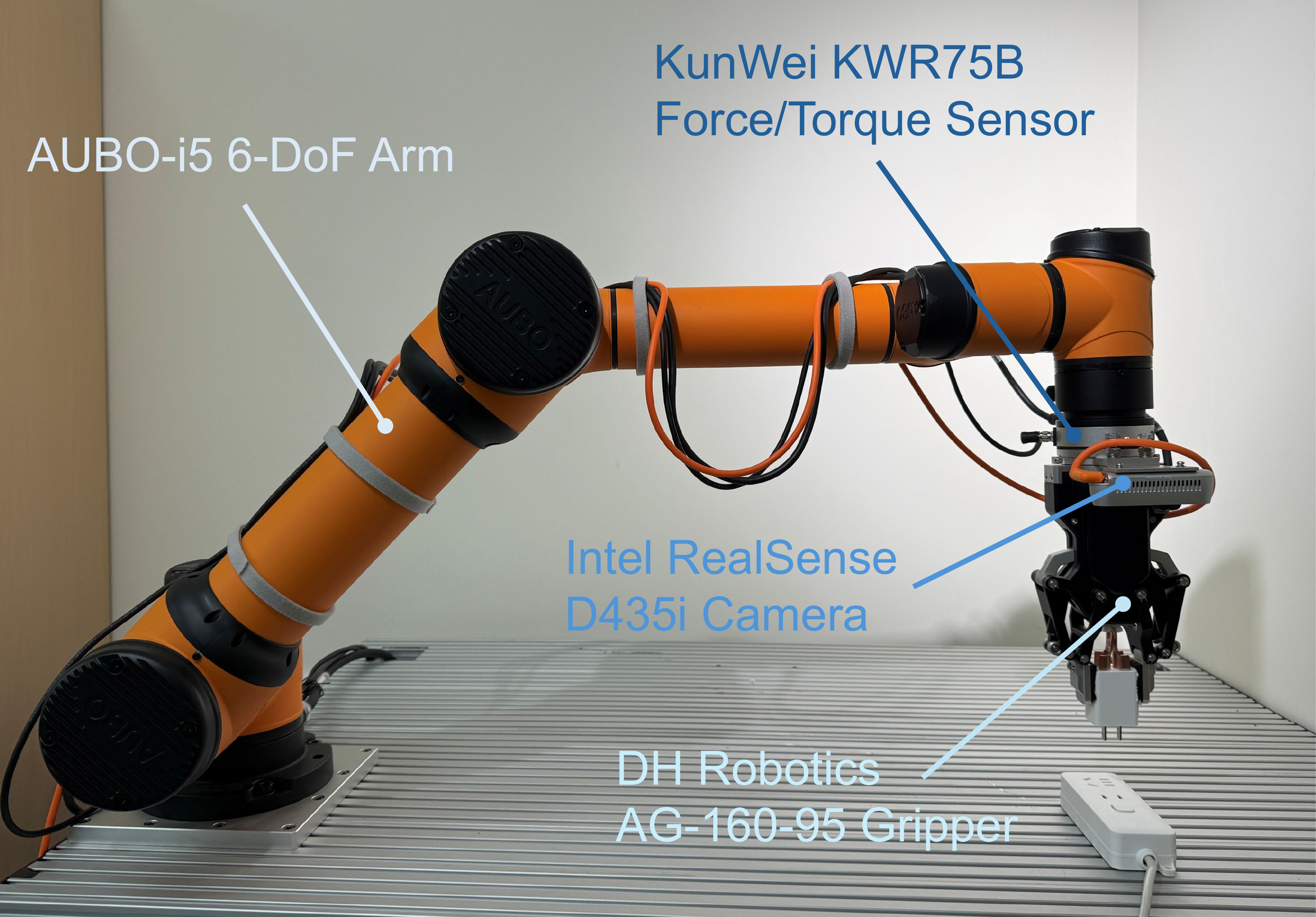}
    \vspace{-2em}
    \caption{\textbf{Real-robot hardware platform.}
    Annotated setup used for the physical connector-insertion study.}
    \label{fig:real-robot-platform-setup}
    \vspace{-1.3em}
\end{figure}

\subsection{Real-Robot Platform}
\label{app:real-robot-platform}

The hardware study uses an AUBO-i5 6-DoF arm, a DH Robotics AG-160-95 gripper held
closed on the connector fixture, a wrist Intel RealSense D435i, an external
D415, and the KunWei KWR75B six-axis force/torque sensor. The connector
protocol uses a 16 mm insertion travel from initial contact to the seated state. The
balanced study uses 90 physical trials, with 30 trials for each of PaCo-VLA, a
nonsemantic shielded controller, and an energy-tank baseline. A separate EV
charging-gun transfer run evaluates PaCo-VLA over 30 trials and completes 17/30
insertions. Included trials retain synchronized external video, wrist
video, RGB-D metadata, F/T traces, proposal traces,
projected-schedule traces, tank-energy traces, and F/T-governor
interventions.

\bibliographystyle{IEEEtran}
\bibliography{references}

\end{document}